 \documentclass[cameraready]{Interspeech}

\title{Automatic Lyric Transcription for Greek Songs:\\
 Scaling and Task Composition Effects in Whisper Adaptation}

\author[affiliation={1}, orcid=0009-0006-2283-5502, correspondingauthor]{Maria}{Frangiadaki}
\author[affiliation={1}, orcid=0009-0003-9865-1433]{Dimitrios}{Damianos}
\author[affiliation={1}, orcid=0000-0003-4513-5040]{Kosmas}{Kritsis}
\author[affiliation={1}, orcid=0000-0002-4185-2344]{Vassilis}{Katsouros}

\address{
    $^1$ Institute for Language and Speech Processing, Athena R.C., Greece
}
\email{\{maria.frangiadaki, d.damianos, kosmas.kritsis, vsk\}@athenarc.gr}


\usepackage{comment}
\usepackage{subcaption}
\usepackage[greek,main=english]{babel}

\begin{document}





\maketitle

\begin{abstract}
Automatic Lyric Transcription (ALT) remains substantially more challenging than speech recognition due to melodic variability, rhythmic irregularity, and accompaniment interference. This is heightened in low-resource languages like Greek, where no prior benchmark for ALT exists. We present the first controlled study of Whisper adaptation for Greek ALT, investigating model scaling effects, task composition via multitask training in transcribe-translate ratios, and two-stage speech-to-singing adaptation. We also curate a segment-level aligned singing dataset based on the Greek Audio Dataset (GAD) using source separation and CTC forced alignment. Results show that scaling consistently improves performance, while multitask learning acts as a beneficial regularizer primarily for smaller-capacity models. The 2-stage adaptation in Whisper Large-v3 achieves a Word Error Rate (WER) of 27.2\%, a significant improvement over zero-shot baselines, establishing the first Greek ALT benchmark.

\end{abstract}

\section{Introduction}

Automatic Speech Recognition (ASR) and recent advances in large-scale multilingual pre-trained models have enabled a broad spectrum of everyday applications. However, ASR systems often degrade when the deployment domain differs from the training distribution. A particularly challenging and underexplored form of domain shift is singing voice. The task of converting sung audio into text, known as Automatic Lyric Transcription (ALT), lies at the intersection of ASR and Music Information Retrieval (MIR) and remains substantially more difficult than conventional speech recognition. Singing differs fundamentally from speech in both acoustic and linguistic structure. Large pitch excursions, sustained vowels, melisma (vowel elongation across multiple notes), rhythmic irregularity, instrumental accompaniment, and expressive articulation violate assumptions learned from speech-dominant corpora \cite{kruspe_MoreWordsAdvancements_2024}. As a result, models trained primarily on spoken language struggle to generalize to sung vocals. 

This acoustic domain shift is even more severe in low-resource languages. While English singing datasets have enabled recent progress in ALT, many languages lack systematically curated singing corpora, aligned annotations, and reproducible evaluation protocols. Greek, despite its rich musical tradition and morphological complexity, has not been studied in a controlled ALT setting. No established benchmark currently exists for Greek singing voice ASR, and the behavior of multilingual models under singing-domain adaptation remains unclear.

This work addresses ALT for Greek singing voice, focusing on the Whisper model  \cite{radford_RobustSpeechRecognition_2022} as a strong multilingual pretrained baseline. We investigate to what extent can adaptation of multilingual pre-trained ASR models such as Whisper improve lyric transcription accuracy, when transferring from speech to singing in a low-resource singing scenario. To answer this question, we develop a complete end-to-end pipeline for Greek ALT and conduct a controlled experimental study, which includes fine-tuning Whisper in various checkpoints under transcription-only training and evaluating multitask and two-stage learning strategies.
The contributions of this work are as follows:

\begin{enumerate}
\item We present the first systematic benchmark for Greek ALT.
\item We curated a fully processed and aligned version of the Greek Audio Dataset (GAD) \cite{makris2014greek}, including source-separated and aligned segmentation, translation pairs, and reproducible train-validation-test splits.
\item We conduct a controlled study of model scale and task composition for singing-domain adaptation in a low-resource language.
\item We introduce a task-pure batching scheme with language-aware pre-processing that stabilizes training on singing voice.
\item We propose a quantitative and qualitative error taxonomy tailored to Greek lyrics, highlighting singing-specific errors.
\item We prove that large-scale transcription-only and 2-stage adaptation is the most effective strategy for Greek ALT, while multitask learning effectively serves as regularization and improves Word Error Rate (WER) results for smaller-capacity models.\footnote{Code, models, and reproducible dataset resources are available at
\url{https://github.com/athena-ilsp/lyrics-transcription} and
\url{https://huggingface.co/collections/ilsp/ilsp-greek-whisper-alt-models},
under Apache 2.0 and CC-BY 4.0 licenses.}
\end{enumerate}

\section{Related Work}

\subsection{ALT} 

Early approaches to ALT relied on conventional ASR pipelines tailored to music, typically employing Hidden Markov Models (HMMs) combined with Gaussian Mixture Models (GMMs) or Deep Neural Networks (DNNs) adapted on singing data \cite{mesaros_AutomaticRecognitionLyrics_2010}. The release of benchmark datasets such as DALI \cite{meseguer-brocal_DALILargeDataset_2018} and DAMP-Sing \cite{dabike_AutomaticLyricTranscription_2019} enabled systematic evaluation, highlighting the persistent acoustic mismatch between speech and singing. The shift towards end-to-end deep learning architectures, such as Connectionist Temporal Classification (CTC) and Attention-based Encoder-Decoder (AED) models, unified acoustic modeling and alignment \cite{stoller_EndtoendLyricsAlignment_2019}. Recent work has also extended these to multimodal setups, demonstrating that auxiliary cues like lip movements or note-level transcriptions can further stabilize decoding in low-SNR conditions \cite{gu_MMALTMultimodalAutomatic_2022, gu_ALT_AMT_2024}. Despite these advances, lyric transcription remains a challenging task that requires robust domain adaptation.

\subsection{Foundation Models and adaptation} 

The advent of large-scale, self-supervised foundation models has redefined the state-of-the-art in speech processing. Architectures like wav2vec 2.0 and its cross-lingual extension, XLS-R, learn general acoustic representations that transfer effectively to singing via fine-tuning \cite{baevski_Wav2vec20Framework_2020, babu_XLSRSelfsupervisedCrosslingual_2021}. More recently, OpenAI's Whisper \cite{radford_RobustSpeechRecognition_2022} has demonstrated remarkable zero-shot robustness due to its massive multilingual pre-training. In the context of MIR, while Whisper exhibits strong performance on clean vocals, its zero-shot accuracy degrades significantly on polyphonic audio \cite{kruspe_MoreWordsAdvancements_2024}. To mitigate this, approaches propose cascading Whisper with Large Language Models (LLMs) for error correction \cite{zhuo_LyricWhizRobustMultilingual_2024}. Further ASR techniques, such as employing parameter-efficient tuning (e.g., LoRA) for scalable specialization \cite{song_LoRAWhisperParameterEfficientExtensible_2024}, or utilizing speech-text joint pretraining (e.g., SpeechLM \cite{zhang_SpeechLMEnhancedSpeech_2023})  further integrate acoustic and linguistic information. However, most existing research focuses on high-resource languages, leaving the efficacy of such foundation models on low-resource singing languages largely unexplored.

Adapting ASR models to low-resource domains often necessitates specialized training strategies. Multitask learning, where the model is jointly optimized on auxiliary tasks such as translation, acts as a regularizer preventing overfitting on small target datasets. In the context of Whisper, the interplay between its transcription and translation tokens offers a unique avenue for multitask adaptation \cite{weiss_SequencetoSequenceModelsCan_2017a}. Additionally, staged fine-tuning strategies have been shown to stabilize ASR training in low-resource scenarios\cite{pillai_MultistageFinetuningStrategies_2024}. Data-centric strategies further bridge the speech-to-singing gap through voice-to-singing augmentation \cite{basak_EndtoendLyricsRecognition_2021} and consistency loss regularization \cite{_EnhancingLyricsTranscription_2025}.

\subsection {Greek Speech and Singing Recognition} 

Automatic Speech Recognition for the Greek language has seen progress through recent spoken corpora that strengthen the ASR infrastructure \cite{paraskevopoulos_GreekPodcastCorpus_2024, vakirtzian_SpeechRecognitionGreek_2024}. Benchmarks demonstrate that while generic models like Whisper \cite{radford_RobustSpeechRecognition_2022} perform sufficiently in spoken Greek, they struggle with dialectal variations and fast-paced articulation. To overcome data scarcity, recent frameworks leverage unsupervised domain adaptation. For example, M2DS2 \cite{paraskevopoulos_SampleEfficientUnsupervisedDomain_2022} and MSDA \cite{_MSDACombiningPseudolabeling_2025} combine self-supervised pre-training with pseudo-label-based teacher-student training to effectively reduce domain mismatch in Modern Greek ASR. The intersection of Greek ASR and singing voice analysis is virtually non-existent in the literature. To date, there is no standardized benchmark for Greek ALT, and no study has systematically evaluated the transferability of multilingual foundation models to Greek singing. While existing Greek music datasets, such as Lyra \cite{papaioannou_LyraDataset_2022}, GAD \cite{makris2014greek} and the Greek Music Dataset (GMD) \cite{makris_GreekMusicDataset_2015}, provide valuable acoustic resources for general Music Information Retrieval (MIR) tasks, they lack the segment-level aligned text annotations required for end-to-end audio-to-lyrics transcription. This work bridges this gap by providing the first controlled evaluation of Whisper on Greek singing voice, establishing a baseline for future research in low-resource ALT.

\section{The GAD-ALT Dataset}
\label{sec:dataset}

A core contribution of this work is the curation of the GAD-ALT corpus. ALT requires temporal synchronization between audio and text, so we extend the GAD \cite{makris2014greek}, which was originally designed for genre classification, into an ASR-ready dataset. The GAD is a collection of 1,000 popular Greek songs spanning multiple genres (Urban,
Folk, Rock, Hip Hop Pop). All entries are accompanied by genre annotations, lyrics, manually annotated mood labels, metadata, extracted audio features and links to the corresponding songs on YouTube, thus enabling researchers to obtain the raw audio when required.

\subsection{Source Separation}
To convert this into a reproducible ALT benchmark, substantial curation was required to resolve missing metadata, correct mismatches, and standardize lyric formatting. Because music recordings are polyphonic, we employ Hybrid Transformer Demucs (\texttt{htdemucs\_ft}) \cite{rouard_HybridTransformersMusic_2022} for two-stem source separation,  extracting vocal and accompaniment tracks. Extracted vocals are downmixed to mono, resampled to 16 kHz to match Whisper's \cite{radford_RobustSpeechRecognition_2022} input requirements, and converted into Kaldi format. 

\subsection{Forced Alignment and Bilingual Augmentation} 

For temporal alignment, we employ a customized version of the open-source CTC forced aligner \cite{ctc_forced_aligner_repo}. Recordings are processed in 30-second overlapping windows. To filter out low-quality alignments, we compute a custom confidence score that combines aligned token percentage (50\%), average CTC log-probabilities (30\%) and duration regularity penalties (20\%) \cite{kurzinger_CTCSegmentationLargeCorpora_2020}. To enable multitask experiments, each aligned Greek segment is translated into English at the segment level using the gpt-4o-mini model via the OpenAI API \cite{openai2023gpt4}, with zero-shot prompting,temperature set to 0 and a maximum limit of 128 tokens, preserving temporal alignment. The final curated singing corpus is in Hugging Face form and comprises 17,458 aligned lyric segments (19.65 hours). To prevent data leakage, splitting is performed at the song level, partitioning the dataset into 13,750 training (78.8\%), 1,892 validation (10.8\%), and 1,816 test segments (10.4\%), with a mean duration of 4 seconds. Each final entry contains: (i) the aligned audio segment, (ii) the normalized Greek transcription, and (iii) the English translation.

\begin{figure}[t]
    \centering
    \includegraphics[width=0.85\columnwidth]{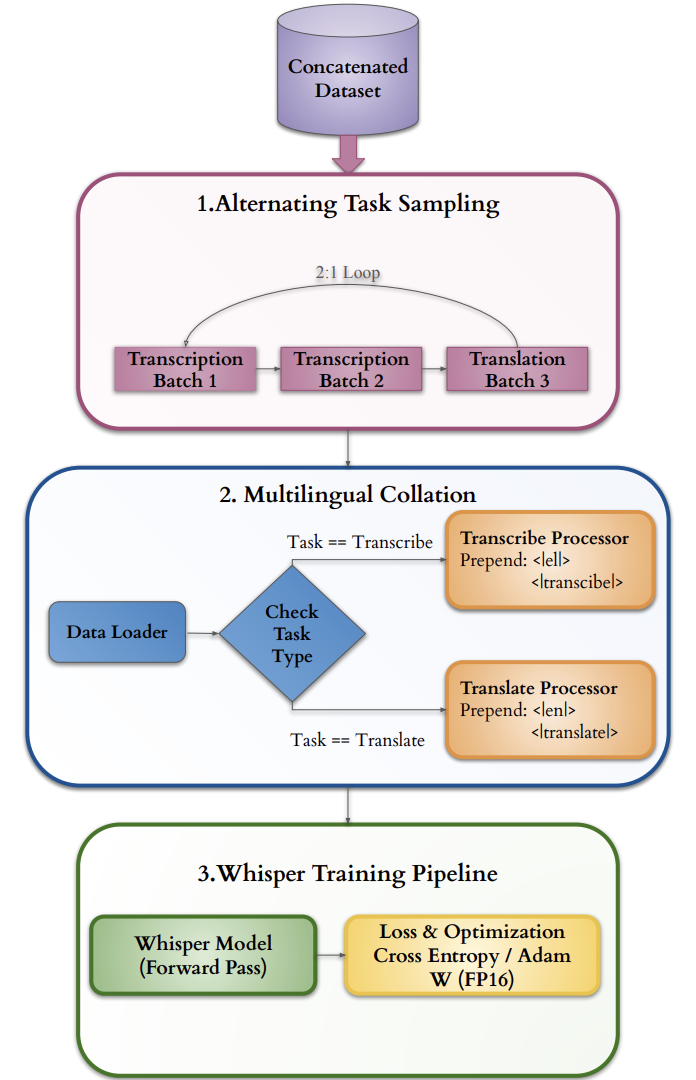}
    \caption{Overview of the proposed multitask training framework for the 2:1 transcription:translation ratio.}
    \label{fig:2to1_methodology}
    \vspace{-0.5cm}
\end{figure}

\section{Methodology and Experimental Setup}

\subsection{Overview} 

The proposed methodology introduces a complete end-to-end pipeline for Greek ALT. After curating the GAD-ALT dataset, multilingual pretrained Whisper models \cite{radford_RobustSpeechRecognition_2022} are adapted to the singing domain under controlled settings that vary in model scale, task composition, and staged speech-to-singing adaptation. 


\subsection{Whisper Adaptation Strategies} 

Zero-shot inference using the pretrained models serves as our out-of-domain baseline. We evaluate three model scales (Small, Medium, Large-v3) under two supervised fine-tuning regimes. A Transcription-only task (Greek audio $\rightarrow$ Greek text) and a Multitask setting (Greek audio $\rightarrow$ Greek transcription + English translation). For the latter, task-specific batches are interleaved in fixed ratios (2:1, 4:1) using a deterministic sampler to ensure task-homogeneous batches, allowing us to investigate if translation acts as a regularizer. The proposed multitask training framework is illustrated in Figure ~\ref{fig:2to1_methodology}. To reduce the speech-to-singing domain gap, we also evaluate a Staged Adaptation strategy. For this, we additionally utilize a Greek subset of Mozilla Common Voice (v23.0) \cite{ardila_CommonVoiceMassivelyMultilingual_2020}, which comprises approximately 37.5 hours of validated read speech. In Stage 1, we fine-tune Whisper solely on this Greek speech while keeping the entire encoder frozen, so that the updates are implemented exclusively to the decoder for language adaptation. Finally, in Stage 2, the model is fully unfrozen and fine-tuned on the singing corpus.

\subsection{Training Details} 
All models are trained using the Hugging Face \texttt{Seq2SeqTrainer} on multi-GPU NVIDIA A100 nodes. Optimization is performed via AdamW for 5 epochs.  Learning rate of $5\times10^{-5}$  performed better for Whisper Small and Medium, whereas $3\times10^{-5}$ proved more suitable for Whisper Large-v3. Per-GPU batch size ranges from 4 to 8 segments depending on model scale, utilizing mixed-precision (FP16) for efficiency. 
To facilitate future research, our complete training pipeline and model configurations will be released as open-source.

\section{Results}

Evaluation is computed in normalized Word Error Rate (WER), after lowercasing, punctuation removal, and standard text normalization. The WER is computed using the \texttt{jiwer} library over aggregated segment-level predictions.

\begin{table}[t]
\centering
\caption{Normalized Word Error Rate (WER \%) on the Greek singing test set. The table contrasts model scaling against task composition (multitask ratios) and staged adaptation. Best adaptation results per model scale are highlighted in bold.}
\label{tab:wer_results}
\begin{tabular}{@{}lccc@{}}
\toprule
\textbf{Training Setup} & \textbf{Small} & \textbf{Medium} & \textbf{Large-v3} \\
\midrule
Zero-shot & 92.3 & 65.1 & 53.6 \\ 
\midrule
2:1 transcribe-translate  & \textbf{33.6} & 32.3 & 30.7 \\
4:1 transcribe-translate  & 34.9 & 31.6 & 30.2 \\
Transcribe-only & 36.7 & 30.3 & 28.4 \\
\midrule
2 stages transcribe & 36.6 & \textbf{30.1} & \textbf{27.2} \\
\bottomrule
\end{tabular}
\end{table}
\subsection{Domain Gap Analysis} 

Table 1 summarizes normalized WER (\%) across all Whisper model sizes and training configurations on the held-out Greek singing test set. Zero-shot evaluation reveals a clear scaling trend. Whisper Small fails almost completely with 92.3\% WER, Whisper Medium achieves 65.1\%, and Whisper Large reduces error further to 53.6\%. Although increased model capacity partially mitigates degradation, performance remains insufficient for practical ALT. In contrast, supervised fine-tuning reduces WER significantly across larger models, demonstrating that scale alone cannot bridge the speech-to-singing domain gap. Thus, zero shot approaches like Lyricwiz \cite{zhuo_LyricWhizRobustMultilingual_2024} would not be ideal for a low resource language like Greek. While Whisper's large-scale pretraining provides a strong foundation, it does not fully account for melodic prolongation, vowel stretching, rhythmic compression, and altered phoneme realizations characteristic of singing voice. 

\subsection{Capacity and Regularization} 

We systematically evaluated Whisper \cite{radford_RobustSpeechRecognition_2022} across scales and training strategies to establish the first benchmark for Greek ALT. Our findings indicate that multitask learning primarily acts as a regularization mechanism for smaller-capacity models, while larger models benefit more from focused transcription-only adaptation. For Whisper Small (244M parameters), the 2:1 transcribe:translate configuration achieves the best WER performance (33.6\%), outperforming transcription-only training (36.7\%). Increasing transcription dominance to 4:1 slightly degrades performance (34.9\%), yet remains superior to pure transcription. This pattern suggests that moderate auxiliary translation exposure provides beneficial regularization at low capacities, encouraging more stable encoder representations without overwhelming the primary transcription objective. 

\begin{figure}[t]
\centering
    \begin{subfigure}{0.9\linewidth}
        \includegraphics[width=\linewidth]{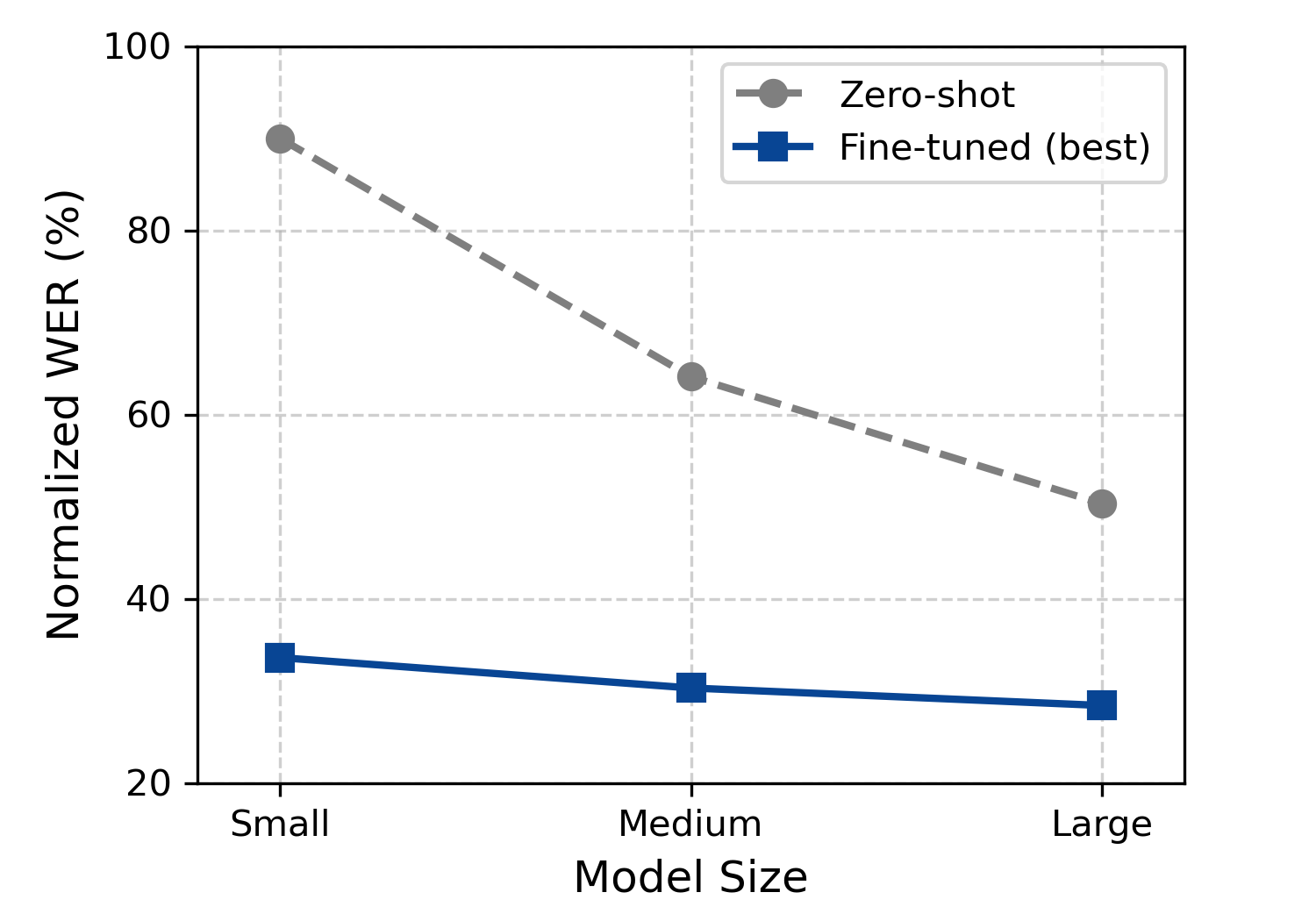}
        \caption*{(a) Scaling effect}
    \end{subfigure}
    
    
    \begin{subfigure}{0.9\linewidth}
        \includegraphics[width=\linewidth]{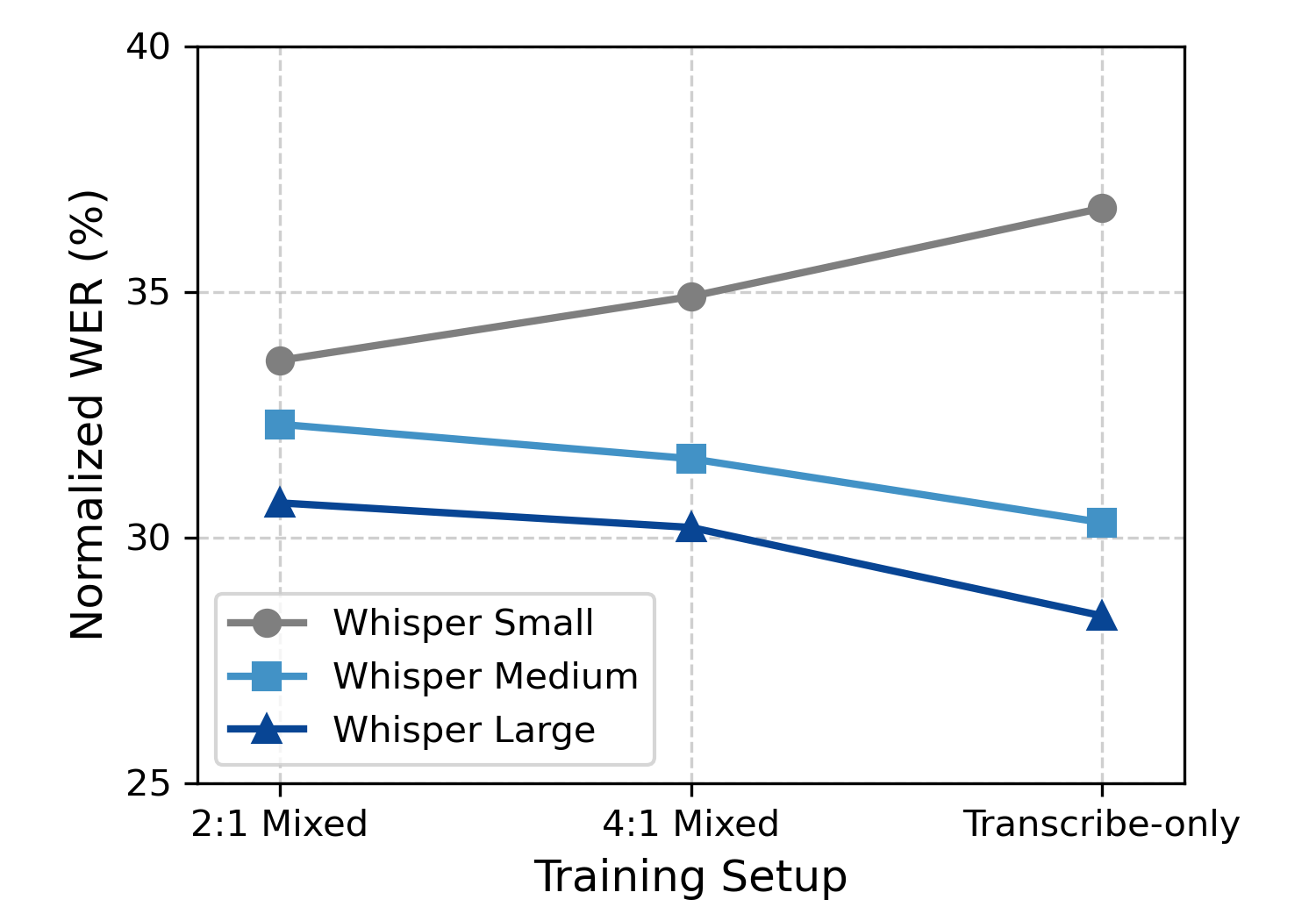}
        \caption*{(b) Multitask ratio effect}
    \end{subfigure}
\caption{Model scaling improves robustness, but singing-domain adaptation dominates performance. Multitask mixing primarily benefits smaller-capacity models.}
\label{fig:scaling_ratio}
\vspace{-0.5cm}
\end{figure}

In contrast, as observed in Figure~\ref{fig:scaling_ratio}, as model capacity increases, the encoder-decoder architecture internalizes sufficient linguistic structure from pretraining and benefits more from target-focused specialization. For Whisper Medium (769M parameters), transcription-only fine-tuning yields the lowest WER (30.3\%), outperforming both 2:1 (32.3\%) and 4:1 (31.6\%) mixtures. A similar trend is observed for Whisper Large-v3 (1.55B parameters), where transcription-only training achieves 28.4\% WER. 

\subsection{Two-Stage Adaptation} 

Interestingly, our two-stage approach proved beneficial for the larger models. The Whisper Small performed better when fine-tuned directly on the singing data. With the 2-stage approach, it achieved 36.6\% WER, a slightly better score than the transcribe-only fine-tune (36.7\%), but still worse than the multitask training (33.6\%). This may also be attributed to the characteristics of the intermediate speech corpus. The Greek Common Voice dataset \cite{ardila_CommonVoiceMassivelyMultilingual_2020} primarily consists of read speech with controlled prosody. As a result, Stage-1 adaptation likely reinforces clean speech patterns that remain acoustically distant from singing. Future work will investigate whether spontaneous, prosodically rich speech corpora (e.g., conversational or podcast-style speech) provide a more acoustically compatible intermediate domain. On the other hand, Whisper medium shows marginal gains from this approach (30.1\% WER), while the Large-v3 model seems to have enough parameters to learn strong Greek language patterns from the speech data without losing the flexibility needed to adapt to the complex acoustics of songs later on, ultimately achieving  the best performance of 27.2\%  WER.

\subsection{Ablations of Source Separation and Augmentation} 

Interestingly, training and evaluating on isolated vocal stems yield measurable gains over using raw polyphonic mixtures. We fine-tuned and tested the model with the best WER score on raw  polyphonic data, and it scored 33.4\% WER . This suggests that residual accompaniment is not the primary error driver in this dataset, but vocal isolation is still a better choice. Furthermore, augmentation strategies based on SNR-controlled stem remixing, light reverberation, and mixed raw+vocals training consistently degraded performance, notably increasing WER compared to the vocal-only baseline. We therefore conclude that vocals-only training without artificial remixing is the most reliable configuration.

\subsection{Qualitative Error Analysis} 

Beyond WER scores, we perform a structured qualitative analysis to characterize singing-specific error patterns in Greek. We manually inspected and categorized 200 transcription errors produced by the best-performing model.
\begin{itemize}
    \item \textbf{Semantic substitution (25.5\%):} The model occasionally replaces a word with a phonetically similar but semantically distinct alternative (e.g., ``\textgreek{χέρι}'' [hand] transcribed as ``\textgreek{γέροι}'' [old men], or ``\textgreek{κρίμα}'' [pity] as ``\textgreek{χρήμα}'' [money]). 
    \item \textbf{Boundary drift (24.0\%):} Melismatic stretching often leads to incorrect segmentation, merging or splitting lexical units (e.g., ``\textgreek{στην αμμουδιά ποτέ του}'' boundaries shifting to create the non-words ``\textgreek{στην αμμου διαποτετου}''), indicating difficulty aligning syllabic timing with word boundaries. 
    \item \textbf{Hallucinated or severely corrupted content (19.5\%):} High rhythmic density and rapid articulation frequently blur consonant clusters, resulting in phonotactically implausible syllables (e.g., a nonsensical string ``\textgreek{κοντεριακος ατρειατα πατασιαζουν της}''). 
    \item \textbf{Orthographic ambiguity (17.0\%):} Greek contains many homophones as well as multiple graphemic representations for similar vowel sounds (e.g., \textgreek{ι/η/ει/υ/οι} all stand for "i", and \textgreek{ο/ω} are both pronounced "o"). Whisper transcriptions contain substitutions that preserve phonetic similarity but alter lexical meaning or grammar (e.g., ``\textgreek{όλοι}'' [all, masculine plural] transcribed as ``\textgreek{όλη}'' [all, feminine singular], or ``\textgreek{σειρήνες}'' [sirens] as ``\textgreek{συρίνες}'' [spelling mistake]). 
    \item \textbf{Function-word deletion and insertion (12.0\%):} Short grammatical particles (e.g., ``\textgreek{μη}'', ``\textgreek{πως}'', ``\textgreek{δε}'') are often omitted in fast singing or spuriously inserted. 
    \item \textbf{Morphological drift (2.0\%):} In several instances, the lemma is preserved but the inflection changes (e.g., the neuter adjective ``\textgreek{πανάκριβο}'' altering its suffix to plural ``\textgreek{πανάκριβα}'').
\end{itemize}
Overall, fine-tuning substantially reduces error frequency but does not eliminate singing-specific categories. Larger models primarily decrease severity rather than altering the distribution of error types, indicating that melodic variability and articulation distortions remain central challenges for Greek ALT.

\section{Conclusions \& Future Work}

In this work, we presented the first systematic benchmark for Greek ALT. We expand the GAD \cite{makris2014greek} to GAD-ALT, with segments, alignment, translations and Hugging Face splits. Through a comprehensive evaluation of Whisper \cite{radford_RobustSpeechRecognition_2022} adaptation strategies, we demonstrated that while zero-shot inference suffers from a severe speech-to-singing domain gap, targeted fine-tuning dramatically reduces the Word Error Rate to 27.2\%. Furthermore, our findings reveal that multitask learning (incorporating translation) acts as an effective regularizer for smaller-capacity models, whereas larger models benefit most from focused, transcription-only and two-stage adaptation. Future work will focus on expanding the curated singing corpus, exploring expressive and spontaneous speech corpora for more effective staged adaptation, using parameter-efficient fine-tuning methods, and integrating Greek-specific language models to better handle the morphological and rhythmic complexities of the singing voice. 

\section{Acknowledgements}

The authors acknowledge the EuroHPC Joint Undertaking for awarding this project access to the EuroHPC supercomputer LEONARDO, hosted by CINECA (Italy) and the LEONARDO consortium, through a EuroHPC Development Access call (Project No. EUHPC-D27-063). This work received partial funding from the European High-Performance Computing Joint Undertaking (JU) under Grant Agreement No. 101234269 for the Pharos AI Factory project, as well as from the Greek Ministry of Digital Governance and Artificial Intelligence.

\section{Generative AI Use Disclosure}

Portions of this manuscript were refined with the assistance of generative AI tools for language editing and clarity. All experimental design, analysis, and scientific conclusions were developed independently by us.

\bibliographystyle{IEEEtran}
\bibliography{Interspeech2026_ALT}

@inproceedings{_EnhancingLyricsTranscription_2025,
  author = {Jiawen Huang and others},
  title = {Enhancing Lyrics Transcription on Music Mixtures with Consistency Loss},
  booktitle = {Proceedings of Interspeech},
  year = {2025},
  doi = {10.21437/Interspeech.2025-311}
}

@inproceedings{_MSDACombiningPseudolabeling_2025,
  author = {Dimitrios Damianos and others},
  title = {{MSDA}: Combining Pseudo-labeling and Self-Supervision for Unsupervised Domain Adaptation in {ASR}},
  booktitle = {Proceedings of Interspeech},
  year = {2025},
  doi = {10.21437/Interspeech.2025-695}
}

@inproceedings{ardila_CommonVoiceMassivelyMultilingual_2020,
  author = {Rosana Ardila and others},
  title = {Common Voice: A Massively-Multilingual Speech Corpus},
  booktitle = {Proceedings of the 12th Language Resources and Evaluation Conference (LREC)},
  year = {2020},
  pages = {4218--4222}
}

@inproceedings{babu_XLSRSelfsupervisedCrosslingual_2021,
  author = {Arun Babu and others},
  title = {{XLS-R}: Self-supervised Cross-lingual Speech Representation Learning at Scale},
  booktitle = {Proceedings of Interspeech},
  year = {2022}
}

@inproceedings{baevski_Wav2vec20Framework_2020,
  author = {Alexei Baevski and Henry Zhou and Abdelrahman Mohamed and Michael Auli},
  title = {wav2vec 2.0: A Framework for Self-Supervised Learning of Speech Representations},
  booktitle = {Advances in Neural Information Processing Systems (NeurIPS)},
  year = {2020}
}

@inproceedings{basak_EndtoendLyricsRecognition_2021,
  author = {Sakya Basak and Shrutina Agarwal and Sriram Ganapathy and Naoya Takahashi},
  title = {End-to-End Lyrics Recognition with Voice to Singing Style Transfer},
  booktitle = {IEEE International Conference on Acoustics, Speech and Signal Processing (ICASSP)},
  year = {2021}
}

@inproceedings{dabike_AutomaticLyricTranscription_2019,
  author = {G. R. Dabike and J. Barker},
  title = {Automatic Lyric Transcription from Karaoke Vocal Tracks: Resources and a Baseline System},
  booktitle = {Proceedings of Interspeech},
  year = {2019},
  pages = {579--583}
}

@inproceedings{gu_MMALTMultimodalAutomatic_2022,
  author = {Xiangming Gu and Longshen Ou and Danielle Ong and Ye Wang},
  title = {{MM-ALT}: A Multimodal Automatic Lyric Transcription System},
  booktitle = {Proceedings of the 30th ACM International Conference on Multimedia},
  year = {2022},
  pages = {3328--3337},
  doi = {10.1145/3503161.3548411}
}

@inproceedings{kruspe_MoreWordsAdvancements_2024,
  author = {Anna Kruspe},
  title = {More than Words: Advancements and Challenges in Speech Recognition for Singing},
  booktitle = {IEEE International Conference on Acoustics, Speech and Signal Processing (ICASSP)},
  year = {2024}
}

@incollection{kurzinger_CTCSegmentationLargeCorpora_2020,
  author = {Ludwig K{\"u}rzinger and Dominik Winkelbauer and Lujun Li and Tobias Watzel and Gerhard Rigoll},
  title = {{CTC}-Segmentation of Large Corpora for German End-to-end Speech Recognition},
  booktitle = {International Conference on Speech and Computer (SPECOM)},
  volume = {12335},
  pages = {267--278},
  publisher = {Springer},
  year = {2020},
  doi = {10.1007/978-3-030-60276-5_27}
}

@article{mesaros_AutomaticRecognitionLyrics_2010,
  author = {Annamaria Mesaros and Tuomas Virtanen},
  title = {Automatic Recognition of Lyrics in Singing},
  journal = {EURASIP Journal on Audio, Speech, and Music Processing},
  volume = {2010},
  number = {1},
  pages = {1--11},
  year = {2010},
  doi = {10.1155/2010/546047}
}

@inproceedings{meseguer-brocal_DALILargeDataset_2018,
  author = {Gabriel {Meseguer-Brocal} and Alice {Cohen-Hadria} and Geoffroy Peeters},
  title = {{DALI}: A Large Dataset of Synchronized Audio, Lyrics and Notes, Automatically Created Using Teacher-Student Machine Learning Paradigm},
  booktitle = {Proceedings of the International Society for Music Information Retrieval Conference (ISMIR)},
  year = {2018},
  doi = {10.5281/zenodo.1492443}
}

@inproceedings{paraskevopoulos_GreekPodcastCorpus_2024,
  author = {Georgios Paraskevopoulos and Chara Tsoukala and Athanasios Katsamanis and Vassilis Katsouros},
  title = {The Greek Podcast Corpus: Competitive Speech Models for Low-Resourced Languages with Weakly Supervised Data},
  booktitle = {Proceedings of Interspeech},
  year = {2024}
}

@inproceedings{paraskevopoulos_SampleEfficientUnsupervisedDomain_2022,
  author = {Georgios Paraskevopoulos and Theodoros Kouzelis and Georgios Rouvalis and Athanasios Katsamanis and Vassilis Katsouros and Alexandros Potamianos},
  title = {Sample-Efficient Unsupervised Domain Adaptation of Speech Recognition Systems: A Case Study for Modern Greek},
  booktitle = {IEEE International Conference on Acoustics, Speech and Signal Processing (ICASSP)},
  year = {2023}
}

@article{pillai_MultistageFinetuningStrategies_2024,
  author = {Leena G. Pillai and Kavya Manohar and Basil K. Raju and Elizabeth Sherly},
  title = {Multistage Fine-tuning Strategies for Automatic Speech Recognition in Low-resource Languages},
  journal = {arXiv preprint arXiv:2411.04573},
  year = {2024}
}

@inproceedings{radford_RobustSpeechRecognition_2022,
  author = {Alec Radford and others},
  title = {Robust Speech Recognition via Large-Scale Weak Supervision},
  booktitle = {International Conference on Machine Learning (ICML)},
  year = {2023}
}

@inproceedings{rouard_HybridTransformersMusic_2022,
  author = {Simon Rouard and Francisco Massa and Alexandre D{\'e}fossez},
  title = {Hybrid Transformers for Music Source Separation},
  booktitle = {IEEE International Conference on Acoustics, Speech and Signal Processing (ICASSP)},
  year = {2023}
}

@inproceedings{song_LoRAWhisperParameterEfficientExtensible_2024,
  author = {Zheshu Song and others},
  title = {{LoRA-Whisper}: Parameter-Efficient and Extensible Multilingual {ASR}},
  booktitle = {Proceedings of Interspeech},
  year = {2024}
}

@inproceedings{stoller_EndtoendLyricsAlignment_2019,
  author = {Daniel Stoller and Simon Durand and Sebastian Ewert},
  title = {End-to-End Lyrics Alignment for Polyphonic Music Using an Audio-to-Character Recognition Model},
  booktitle = {IEEE International Conference on Acoustics, Speech and Signal Processing (ICASSP)},
  year = {2019},
  pages = {181--185}
}

@inproceedings{vakirtzian_SpeechRecognitionGreek_2024,
  author = {Socrates Vakirtzian and others},
  title = {Speech Recognition for Greek Dialects: A Challenging Benchmark},
  booktitle = {Proceedings of Interspeech},
  year = {2024},
  pages = {3974--3978}
}

@inproceedings{weiss_SequencetoSequenceModelsCan_2017a,
  author = {Ron J. Weiss and Jan Chorowski and Navdeep Jaitly and Yonghui Wu and Zhifeng Chen},
  title = {Sequence-to-Sequence Models Can Directly Translate Foreign Speech},
  booktitle = {Proceedings of Interspeech},
  year = {2017}
}

@article{zhang_SpeechLMEnhancedSpeech_2023,
  author = {Ziqiang Zhang and others},
  title = {{SpeechLM}: Enhanced Speech Pre-Training with Unpaired Textual Data},
  journal = {IEEE/ACM Transactions on Audio, Speech, and Language Processing},
  year = {2023}
}

@inproceedings{zhuo_LyricWhizRobustMultilingual_2024,
  author = {Le Zhuo and others},
  title = {{LyricWhiz}: Robust Multilingual Zero-shot Lyrics Transcription by Whispering to {ChatGPT}},
  booktitle = {IEEE International Conference on Acoustics, Speech and Signal Processing (ICASSP)},
  year = {2024}
}

@article{gu_ALT_AMT_2024,
  author = {Xiangming Gu and others},
  title = {Automatic Lyric Transcription and Automatic Music Transcription from Multimodal Singing},
  journal = {ACM Transactions on Multimedia Computing, Communications and Applications},
  volume = {20},
  number = {7},
  pages = {1--29},
  year = {2024},
  doi = {10.1145/3651310}
}

@inproceedings{makris2014greek,
  title={The Greek Audio Dataset},
  author={Makris, Dimos and Kermanidis, Katia L and Karydis, Ioannis},
  booktitle={IFIP International Conference on Artificial Intelligence Applications and Innovations},
  pages={1--10},
  year={2014},
  organization={Springer}
}

@inproceedings{papaioannou_LyraDataset_2022,
  author = {Charilaos Papaioannou and Ioannis Valiantzas and Theodoros Giannakopoulos and Maximos Kaliakatsos-Papakostas and Alexandros Potamianos},
  title = {A Dataset for Greek Traditional and Folk Music: Lyra},
  booktitle = {Proceedings of the 23rd International Society for Music Information Retrieval Conference (ISMIR)},
  year = {2022},
  pages = {344--351},
  address = {Bengaluru, India}
}

@inproceedings{makris_GreekMusicDataset_2015,
  author = {Dimos Makris and Ioannis Karydis and Spyros Sioutas},
  title = {The Greek Music Dataset},
  booktitle = {Proceedings of the 16th International Conference on Engineering Applications of Neural Networks (EANN)},
  year = {2015},
  pages = {22:1--22:7},
  publisher = {ACM},
  address = {Rhodes, Greece}
}

@misc{ctc_forced_aligner_repo,
  author = {Mahmoud Ashraf},
  title = {CTC Forced Aligner},
  year = {2023},
  publisher = {GitHub},
  journal = {GitHub repository},
  howpublished = {\url{https://github.com/MahmoudAshraf97/ctc-forced-aligner}}
}

@article{openai2023gpt4,
  title={GPT-4 Technical Report},
  author={OpenAI},
  journal={arXiv preprint arXiv:2303.08774},
  year={2023}
}

\end{document}